\documentclass[sigconf]{acmart}

\copyrightyear{2025}
\acmYear{2025}
\setcopyright{cc}
\setcctype{by-nc-nd}
\acmConference[COMPASS '25]{ACM SIGCAS/SIGCHI Conference on Computing
and Sustainable Societies}{July 22--25, 2025}{Toronto, ON, Canada}
\acmBooktitle{ACM SIGCAS/SIGCHI Conference on Computing and Sustainable
Societies (COMPASS '25), July 22--25, 2025, Toronto, ON, Canada}
\acmDOI{10.1145/3715335.3735483}
\acmISBN{979-8-4007-1484-9/2025/07}

\usepackage[utf8]{inputenc} 
\usepackage[T1]{fontenc}    
\usepackage{hyperref}       
\usepackage{url}            
\usepackage{booktabs}       
\usepackage{amsfonts}       
\usepackage{nicefrac}       
\usepackage{microtype}      

\usepackage{graphicx}       
\usepackage{multirow}
\usepackage{array}
\usepackage{colortbl}
\usepackage{xcolor}
\usepackage{amsmath}
\usepackage{subcaption}
\usepackage{makecell}
\usepackage{float}          

\AtBeginDocument{
  \providecommand\BibTeX{{
    \normalfont B\kern-0.5em{\scshape i\kern-0.25em b}\kern-0.8em\TeX}}}

\setcopyright{none}

\newcommand{\Noah}[1]{{\color{black}#1}}

\newcommand{\revise}[1]{{\color{black}#1 }}

\newcommand\blfootnote[1]{
  \begingroup
  \renewcommand\thefootnote{}\footnote{#1}
  \addtocounter{footnote}{-1}
  \endgroup
}

\begin{document}

\title{A Water Efficiency Dataset for African Data Centers}

\author{Noah Shumba}
\affiliation{
  \institution{Carnegie Mellon University Africa}
  \city{Kigali}
  \country{Rwanda}
}

\author{Opelo Tshekiso}
\affiliation{
  \institution{Carnegie Mellon University Africa}
  \city{Kigali}
  \country{Rwanda}
}

\author{Pengfei Li}
\affiliation{
  \institution{Rochester Institute of Technology}
  \city{Rochester}
  \state{New York}
  \country{USA}
}

\author{Giulia Fanti}
\affiliation{
  \institution{Carnegie Mellon University}
  \city{Pittsburgh}
  \state{Pennsylvania}
  \country{USA}
}

\author{Shaolei Ren}
\affiliation{
  \institution{University of California, Riverside}
  \city{Riverside}
  \state{California}
  \country{USA}
}



\begin{abstract}
Artificial intelligence (AI) computing and data centers consume large amounts of freshwater, both directly for cooling
and indirectly for electricity generation. While most attention
has been paid to developed countries such as the U.S.,
this paper presents the first-of-its-kind dataset that
combines nation-level weather and electricity generation data to estimate water usage effectiveness
for data centers in 41 African countries across five different climate regions. 
 We also use our dataset to evaluate and estimate the water consumption of inference on 
two large language models  (i.e., Llama-3-70B and GPT-4) in 11 selected African countries. 
Our estimates suggest that writing a 10-page report using Llama-3-70B could consume as much as \textbf{0.66 liters} of water,
while the water consumption by GPT-4 for the same task may go up to about \textbf{59 liters}. For writing a medium-length email of 120-200 words,
Llama-3-70B and GPT-4 could consume about \textbf{0.13 liters}
and  \textbf{2.9 liters}  of water, respectively.  All the numbers for generative model inference tasks are based on public information available in 2024, when we initially prepared the analysis. Since then, AI inference systems have improved substantially. For example, recent disclosures suggest that energy efficiency improved by more than 30x between May 2024 and May 2025~\cite{DBLP:journals/corr/abs-2508-15734}. Accordingly, our 2024 estimates should be interpreted as historical reference values rather than as representative of current performance.
Interestingly, given the same AI model, 9 of the 11 selected African countries consume less water than the global average, 
mainly because of lower water intensities for electricity generation.
In particular, Libya and Tunisia have an even lower water consumption than the U.S. average, largely due to their reliance on natural gas with low water intensity.
 However, Ethiopia and the Republic of the Congo, whose electricity grids rely almost entirely on hydroelectric generation, exhibit substantially higher water consumption than the global average due to high reservoir evaporation. Our dataset is publicly available.\footnote{ \url{https://huggingface.co/datasets/PengfeiLi/WaterEfficientDatasetForAfricanDataCenters}}
\end{abstract}
\maketitle

\blfootnote{\Noah{This version of the manuscript corrects an earlier minor data processing error that affected some of our onsite water usage effectiveness (WUE) numbers. Because total water use is dominated by offsite water consumption, our overall findings remain similar, although the rankings of some countries have shifted. All other data, including the energy estimates described in Section 4, remain unchanged and were based on public sources available as of mid-2024, when the original manuscript was written.}}

\section{Introduction}
With the rapid growth of artificial intelligence (AI) and digital services, the demand for data centers has increased  substantially~\cite{data_center_energy_growth_2020_McKinsey}. While data center infrastructure 
was historically lacking in Africa, the continent's burgeoning digital economy
has recently led to a surge in data center constructions, with a projected market growth of 50\% by 2026 compared to 2021 \cite{DataCenter_Africa_Growth_50_2024}.

Data centers are notorious for their massive  energy 
usage and water consumption, which have raised significant concerns
even in developed countries such as the U.S. ~\cite{ahmed2014can,DataCenter_US_EnergyUsageReport_2016_LBNL,li2023making}. 
More critically, the added pressure on local water resources
is particularly acute in Africa, where many countries are already grappling with extended droughts and water scarcity challenges \cite{unicef,water-africa}.  
Therefore, it is important to assess data centers' water consumption in Africa, supporting
healthy development of the data center industry for essential economic growth while
ensuring responsible utilization of limited freshwater resources.
While recent studies have begun to address
the growing water consumption of data centers and AI computing \cite{gupta2024dataset,Shaolei_Water_SpatioTemporal_GLB_TCC_2018_7420641,Water_FlexCool_DataCenter_Energy_Water_Tradeoff_AndrewChien_France_eEnergy_2024_10.1145/3632775.3661936,li2023making}, they have predominantly focused
on regions with large data center concentrations, such as the U.S.
and Europe, while leaving out Africa---despite its rapid expansion of data centers and  pressing challenges of water scarcity 
\cite{statista2025datacenter,who2025water}.

In this paper, we address the critical gap in the literature and present a first-of-its-kind water efficiency dataset 
 for data centers in 41 African countries across five distinct climate regions. The dataset includes hourly estimates of water usage effectiveness (WUE)
for both direct and indirect water consumption 
over one year. We obtain these estimates by combining weather data from across Africa with the corresponding fuel mix data (i.e., the composition of energy sources  in each country).
Unlike prior work  \cite{shumba2024waterefficiencydatasetafrican}, we incorporate water stress levels and infrastrcuture inefficiences such as leakage. 
To the best of our knowledge, our work is the first study on AI sustainability to analyze these factors, which strongly influence water consumption. 

To demonstrate the utility of this dataset, we estimate the water consumption of two recent large language models (LLMs), i.e., Llama-3-70B and GPT-4, across 11 selected African countries and compare their AI inference water consumption with that in the U.S. and globally. 
We also illustrate how these estimates compare to water scarcity in various countries.

These insights can be important for policy development, data center site selection,\textbf{ }and infrastructure planning. By making our dataset publicly available, we aim to inform sustainable AI deployment strategies that align with Africa’s unique environmental and resource constraints.

\section{Background and Methodology}
\label{sec:background}
While our dataset is partly based on the methodology and modeling of \cite{gupta2024dataset,li2023making}, which study WUE and AI model water consumption with a heavy emphasis on the U.S. data centers, we also incorporate regional water stresses and infrastructure inefficiencies due to leakages in our model.

\subsection{Water Usage Effectiveness (WUE)}
WUE is measured in units of L/kWh: liters of water consumed per kWh of energy used. 
Like \cite{gupta2024dataset}, we do not model supply chain manufacturing because this aspect often relies on generalized, less accurate data that may not reflect the unique operational practices of individual data centers or computing workloads.
The methodology in \cite{gupta2024dataset} provides
equations for modeling \emph{onsite WUE}, which refers to water directly consumed/evaporated to cool down the facility for each unit of server energy consumption, and  \emph{offsite WUE}, which is also
called the electric water intensity factor and refers to indirect water consumption by the generation of electricity that
supplies each unit of data center energy.
 Note that water \emph{consumption} is defined as 
``water withdrawal minus water discharge'', i.e., the evaporated
portion of water withdrawal that may not be immediately available for reuse
 \cite{reig2020guidance}. Data centers commonly consume 80\% of their direct
 freshwater withdrawal (in many cases, potable water), while only about 10\% of the water withdrawal is consumed by
typical households and offices \cite{Google_SustainabilityReport_2024}.

\subsubsection{Onsite Water Usage Effectiveness} 
To assess onsite WUE, \cite{gupta2024dataset} presents an empirical model created from a commercial cooling tower by considering two configurations.
The first configuration is called ``\emph{fixed approach}'', which fixes the differential between wet-bulb and cold water temperatures, and the second one
``\emph{fixed cold water temperature}'' sets a constant cold water temperature. 
The WUE formulas for these two configurations are as follows:
\begin{eqnarray}
&\gamma_{\text{Approach}} = \left[-0.0001896 \cdot T_w^2 + 0.03095 \cdot T_w + 0.4442\right]^+,&
\label{eq:1}\\
&\gamma_{\text{ColdWater}} = \left[0.0005112 \cdot T_w^2 - 0.04982 \cdot T_w + 2.387\right]^+,&
\label{eq:2}
\end{eqnarray} 
where $T_w$ is the wet-bulb temperature in Fahrenheit and
$\left[x\right]^+=\max\{0,x\}$. Unless otherwise noted, we will focus on Equation~\eqref{eq:2} and simply refer it to as onsite WUE (i.e., $\gamma_{\text{off}}=\gamma_{\text{ColdWater}}$), because it is typically easier to set a fixed
cold water temperature without adjustment in real systems.
While the onsite WUE for a cooling tower can differ 
from other cooling methods such as air economization with water
evaporation, we note that cooling towers are
one of the most commonly adopted and efficient heat rejection mechanisms for data centers \cite{Equinix_EnvironmentalSustainabilityReport_2024,Google_SustainabilityReport_2024}, especially in hot regions like Africa.

\subsubsection{Offsite  Water Usage Effectiveness} 
Electricity generation is water-intensive and must respond to demand in real-time to maintain grid stability.
Thus, similar to 
 carbon emissions associated with electricity usage,
data centers are also accountable for their electricity water consumption.
Technology companies
 (e.g., Meta) have recently begun to include indirect water consumption for electricity generation in their sustainability reports \cite{Facebook_SustainabilityReport_2024}.
This is critical for holistically understanding the true water impact of data centers, especially in regions where the energy mix includes significant hydroelectric and/or thermal power generation with high water intensities \cite{reig2020guidance,Water_DataCenterFootprint_EnvironmentalResearcHLetters_VT_2021_siddik2021environmental}. Based on \cite{gupta2024dataset,reig2020guidance}, we use the following offsite WUE formula:
\begin{equation}
\gamma_{\text{off}}(t) = \frac{\sum_{k}e_{k}(t) \cdot w_{k}}{\sum_{k}e_{k}(t)},
\label{eq:3}
\end{equation}
where $e_k(t)$ is the amount of electricity produced by energy fuel type $k$ (e.g., hydroelectric, geothermal, coal) at time $t$ and $w_k$ is the corresponding water consumption or intensity factor in L/kWh.

\subsection{Water Leakage}
Leakage refers to water lost during transmission to its destination \cite{10.2166/ws.2024.071}; in our case, the destination of interest is data centers. 
Existing studies quantifying data center water usage in the U.S. and Western regions typically do not model the impact of water infrastructure damage \cite{Shaolei_Water_AI_Thirsty_CACM,Shaolei_Thirsty_DataCenter_Tech,gupta2024dataset}. 

Incorporating leakage into water usage estimates is crucial for regions with aging or poorly maintained infrastructure. 
In many African countries, deteriorating water infrastructure contributes to substantial losses. According to \cite{10.2166/ws.2024.071}, global water leakage rates average around 39\%, while Africa-specific leakage is estimated at 46\%. Some countries, such as Nigeria, experience extreme losses exceeding 61\%, significantly affecting the effective water availability for data center operations. 

To account for this issue, we account for water leakage in our estimates, modeling it as a proportional loss of transmitted water for onsite cooling. Using {country-level average} leakage estimates from \cite{ibnet_non_revenue_water}, we adjust onsite water consumption estimates. Specifically, to estimate $\mathcal L$, the rate of water lost in transmission per kWh of server energy used, we multiply the baseline onsite WUE by the leakage rate $\mathcal \ell$ for a given country: 
\begin{eqnarray}
    \mathcal L = \ell \cdot \gamma_{\text{on}},
\end{eqnarray}
where the leakage rate $\ell$ is taken directly from \cite{ibnet_non_revenue_water}.
This lost volume is then added to the baseline WUE to obtain the adjusted onsite water consumption:
\begin{eqnarray}
    \tilde \gamma_{\text{on}} = \gamma_{\text{on}} + \mathcal L, 
\end{eqnarray}
and our subsequent analysis uses these leakage-adjusted rates $\tilde \gamma_{\text{on}}$.

\subsection{Water Stress}
The impact of data centers' water usage varies heavily with the level of \emph{water stress} in a country, which roughly measures how much water is required for a nation's needs, as a function of water supply. Higher water stress levels indicate greater competition for limited water supplies, making it more challenging to justify water-intensive applications, including AI-driven data centers.

To assess this impact, we compared our water consumption estimates with Baseline Water Stress (BWS) values across different countries \cite{luo2015aqueduct}. BWS is defined as the ratio of total annual water withdrawals to available renewable water supply, providing a standardized measure of water scarcity. The BWS scale ranges from 0 to 5, where:  0–1 represents low stress, 1–2 low to medium stress, 2–3 medium to high stress, 3–4 high stress, and 4–5 extremely high stress. 

This study utilizes total water stress values \cite{luo2015aqueduct}, which aggregate withdrawals across domestic, industrial, and agricultural sectors. This approach captures both direct water consumption and indirect water usage from electricity generation.

\section{Data Collection}
We describe how we collect the necessary data to compute onsite and offsite WUEs, respectively.

\setlength{\textfloatsep}{10pt}
\setlength{\floatsep}{10pt}

\begin{table}[ht]
\centering
\caption{Climate regions and representative countries}
\label{tab:climatic_regions}

\begin{tabular}{ll}
\toprule
Climate Region        & Representative Countries      \\ \midrule
Rainforest             & Republic of the Congo, Gabon, Rwanda \\
Savanna                & Morocco, Tunisia              \\
Desert                 & Egypt, Libya                  \\
Steppe                 & Namibia, Ethiopia             \\
Mediterranean          & Algeria, South Africa         \\ \bottomrule
\end{tabular}

\end{table}

\begin{itemize}
\item \textbf{Weather data.}
For weather data, we first identify five distinct climate regions
in Africa: \emph{Rainforest, Savanna, Desert, Steppe, and Mediterranean regions} \cite{africa2024}.
We then collect weather data from the
countries for each climate region, consisting of hourly wet-bulb temperature, humidity, and precipitation
over one year
from August 23, 2023 to August 22, 2024.
All the weather data are obtained from WeatherAPI \cite{weatherapi2024}, which collects data via ground-based weather stations and satellite imagery.
We then pick the high and low extremes in terms of the average wet-bulb temperature for each region to obtain a representative range.
The selected 11 representative countries are summarized in Table~\ref{tab:climatic_regions}.

\item \textbf{Energy fuel mix.}
We collect the energy fuel mix (i.e., the composition of energy fuels) for electricity generation
in each selected country sourced from OurWorldInData \cite{owid-energy-mix}.
Due to the lack of access to fine-grained data, we use
annual granularity for estimating the offsite WUE as done in the prior literature \cite{reig2020guidance}.
Additionally, we need the water intensity of each fuel type in each selected country to compute offsite WUE in \eqref{eq:3}.
While direct data on the water consumption of various energy fuel types for African countries is lacking, \cite{sanchez2020freshwater} studies water withdrawal and consumption throughout different stages of energy production in Africa. Thus,
we use \cite{sanchez2020freshwater} to derive the average water intensity for each energy fuel type in Africa.

\Noah{Table~\ref{tab:wk_values} lists the African and global water intensity estimates adopted in this work; the global estimates are taken from \cite{reig2020guidance,sanchez2020freshwater}.}

\Noah{Since \cite{sanchez2020freshwater} reports water consumption factors for multiple cooling systems and technology types per fuel (Table~3 therein), we must select a representative cooling technology for each fuel. For most countries, we use cooling tower (CT)-based consumption factors as a conservative baseline. We select the water consumption factors ($w_k$) in Table~\ref{tab:wk_values} that best align with African deployments of power technologies~\cite{sanchez2020freshwater}, including combined-cycle gas turbine (CCGT) for gas and subcritical coal
(0.042~L/kWh)~\cite{sanchez2020freshwater}. For natural gas specifically, we use the combined-cycle rather than subcritical steam value because Africa's major gas-generating countries, including Egypt~\cite{siemens_egypt_ccgt}, Algeria~\cite{powtech_biskra,ge_tm2500_algeria}, Tunisia~\cite{tunisia_rades_ccgt}, and Libya~\cite{libya_benghazi_ccgt} operate predominantly modern CCGT plants. Similarly, we use the subcritical coal value because Africa's coal fleet is predominantly subcritical~\cite{gem_medupi,powtech_kusile}. The exception to CT-based cooling is solar, where we use the photovoltaic (PV) factor of 0.023~L/kWh~\cite{sanchez2020freshwater}, since PV panels involve no thermal cycle or cooling system and the vast majority of Africa's solar capacity is PV, with concentrated solar power (CSP) limited to Morocco and South Africa~\cite{irena_capacity_2023,esi_africa_csp}. Wind (0.001~L/kWh) similarly requires no cooling~\cite{sanchez2020freshwater}. Hydroelectric intensity (5.32~L/kWh) follows the reservoir evaporation methodology in \cite{sanchez2020freshwater} and \cite{mekonnen2015water}.
}

\begin{table*}[t]
\centering
\caption{
\Noah{Water consumption factors ($w_k$) by energy source, technology, and cooling system.}}
\label{tab:wk_values}
\renewcommand{\arraystretch}{1.3}
\begin{tabular*}{\textwidth}{@{\extracolsep{\fill}}lllcc@{}}
\toprule
\textbf{Energy Source} & \textbf{Technology} & \textbf{Cooling} & \textbf{$w_k$ (L/kWh)} & \textbf{$w_k$ (L/kWh)} \\
                       & \textbf{(Africa)}   & \textbf{(Africa)} & \textbf{(Africa)}       & \textbf{(Global)}       \\
\midrule
Coal        & Subcritical steam  & CT   & 2.008  & 2.008  \\
Natural Gas & CCGT               & CT   & 0.795  & 1.580  \\
Oil         & Subcritical steam  & CT   & 2.765  & 2.765  \\
Biomass     & Subcritical steam  & CT   & 2.095  & 2.095  \\
Nuclear     & Subcritical        & OT   & 1.515  & 2.120  \\
Geothermal  & Flash/Binary       & CT   & 0.042  & 0.042  \\
Solar       & Photovoltaic (PV)  & None & 0.023  & 0.400  \\
Wind        & Onshore turbines   & None & 0.001  & 0.001  \\
Hydropower  & Reservoir-based    & Evap.& 5.320  & 22.000 \\
\bottomrule
\end{tabular*}
\end{table*}

\item \textbf{Leakage.}
We obtained national leakage data from \cite{10.2166/ws.2024.071,ibnet_non_revenue_water}, which contains information about global water and sanitation utilities (IBNET).

By providing a standardized framework for collecting and sharing core cost and performance indicators, IBNET enables water utilities to identify leakage hotspots, for instance.

\item \textbf{Water Stress Levels.}
To account for water scarcity and competition, we use baseline water stress (BWS) values  from  \cite{luo2015aqueduct}, a platform for assessing global water risks.

\end{itemize}

\begin{figure}[t]
\centering
\includegraphics[width=0.7\linewidth]{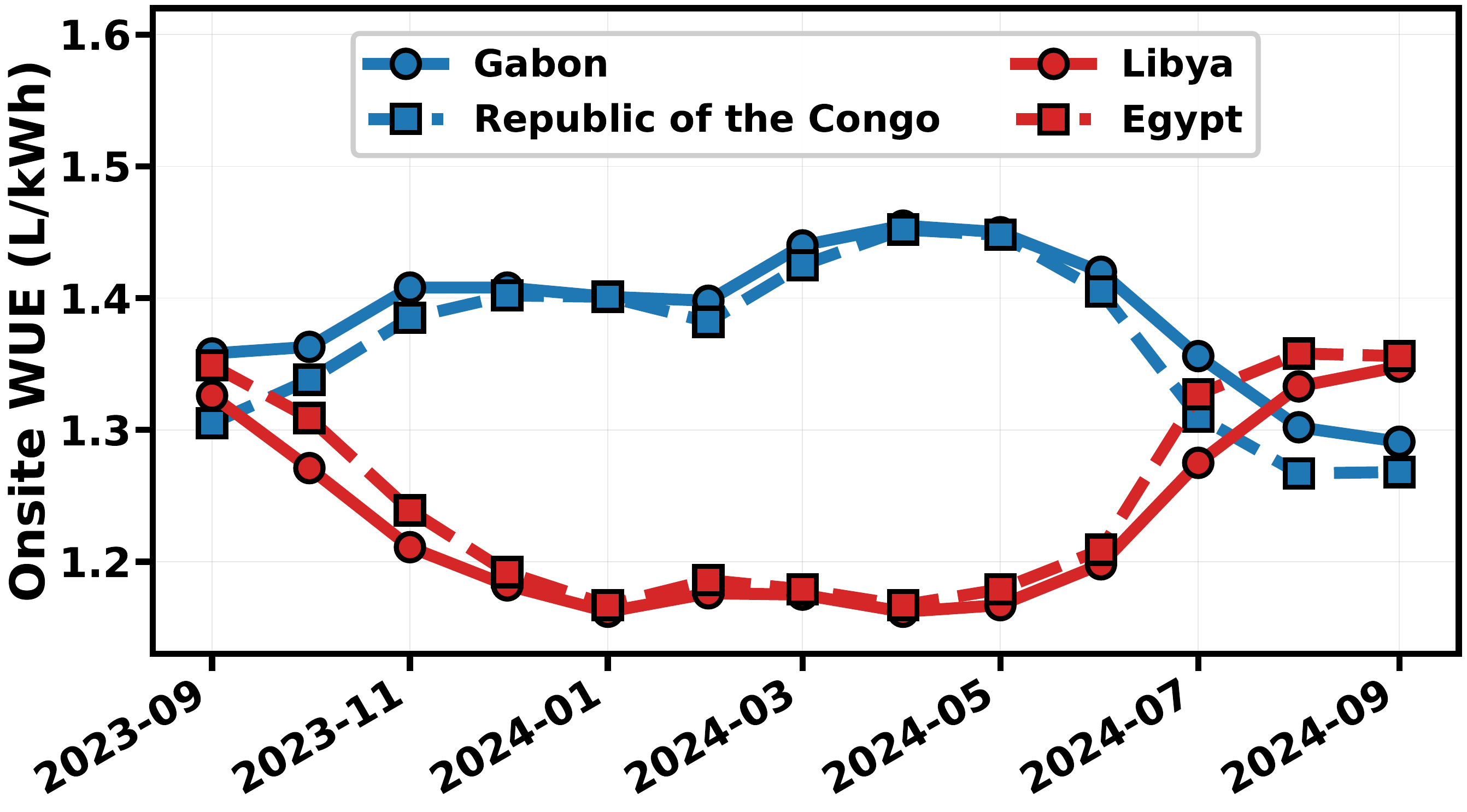}
\caption{Monthly average baseline onsite WUE 
for desert (red) and rainforest (blue) regions.} 
\label{fig:average_onsite_WUE}

\Description{}
\end{figure}

\section{Dataset Evaluation}
Our final dataset provides onsite and offsite (hourly) WUE for capital cities in 41 African countries, as well as leakage. 
For clarity, Figure~\ref{fig:average_onsite_WUE} illustrates the monthly averages for a few selected countries in the rainforest and desert regions. The plot clearly illustrates seasonal trends, as well as onsite WUE differences of \Noah{about 12\%} between climate regions. \Noah{Specifically, rainforest regions generally exhibit higher onsite water consumption than desert regions, as their warmer wet-bulb temperatures require more cooling tower evaporation to reach a target temperature.}

One downstream use for our dataset is to estimate the cost of LLM inference in different countries. 
We first describe our methodology for computing these estimates, followed by our results. 

\subsection{Methodology: Estimating Water Consumption of LLM Inference}\label{Inferancing}

We also present the details of estimating water consumption
by two LLMs, i.e., Meta's Llama-3-70
B and OpenAI's GPT-4.
We use the following equations to calculate the onsite and offsite water consumption, respectively:
\begin{eqnarray}\nonumber
W_{\text{on}} =  \gamma_{\text{on}}\cdot E\;\;\;\;{ \text{and} }\;\;\;\;W_{\text{off}} =  \gamma_{\text{off}}\cdot \rho\cdot E,
\label{eq:4}
\end{eqnarray}
where $W$ is the water consumption,
$\gamma$ is the WUE, $\rho$ is the power usage effectiveness (PUE),
$E$ is the server energy consumption for AI models,
and the subscript ``$\text{on}$'' and ``$\text{off}$'' denote
``onsite'' and ``offsite'' wherever applicable, respectively.
To incorporate leakage, we also compute the adjusted onsite WUE:
\begin{eqnarray}\nonumber
\tilde W_{\text{on}} =  \tilde \gamma_{\text{on}}\cdot E.
\end{eqnarray}

Thus, to estimate the LLMs' water consumption, we need
their  onsite and offsite WUEs (including the onsite WUE adjusted for leakage), energy consumption, and PUE. 

\subsubsection{WUE}
For African countries, we use the average onsite and offsite WUEs from our dataset.  For the U.S. and global references, 
their average onsite WUEs are obtained from publicly accessible reports
based on Microsoft's U.S. average (0.55 L/kWh) and Equinix's global average (1.07 L/kWh), which
represent efficient hyperscale and colocation
data centers, respectively \cite{equinixPUE,Microsoft_Water_2022_how_azure_cloud}. 
Their average offsite WUEs are acquired from the World Resource Institute report \cite{reig2020guidance}.

\subsubsection{Power Usage Effectiveness}
Power Usage Effectiveness (PUE) is a metric that assesses the energy efficiency of a data center by comparing the total energy consumed by the facility to the energy used by the computing equipment. The ideal PUE is 1.0, indicating 100\% energy efficiency in computing. The inference energy estimate 
provided by \cite{ecologits-calculator} assumes a default PUE of 1.2.
 The PUE overhead is not needed for calculating the onsite water consumption, but
should be considered when assessing the offsite water consumption.
For different African countries, we consider an average country-/region-wise PUE 
 
provided by
\cite{DataCenter_PUE_Africa_172_2021}. By taking
the lowest when multiple values are presented in \cite{DataCenter_PUE_Africa_172_2021}, the PUE values for the
11 selected African countries
are as follows:

\begin{table}[H]
    \centering
    \begin{tabular}{l|lll|l} 
    \cmidrule(r){1-2} \cmidrule(r){4-5}
    \textbf{Country} & \textbf{PUE}  &  & \textbf{Country} & \textbf{PUE}  \\ 
    \cline{1-2}\cline{4-5}
    Algeria  & 2.3 &  & Namibia               & 2.1  \\
    Egypt    & 2.3 &  & Republic of the Congo & 2    \\
    Ethiopia & 1.5 &  & South Africa          & 1.4  \\
    Gabon    & 1.9 &  & Tunisia               & 2.3  \\
    Libya    & 2.3 &  & Rwanda                & 1.7  \\
    \cline{4-5}
    Morocco  & 2.3                           \\
    \cmidrule(r){1-2}
    \end{tabular}
    \vspace{0.2cm}
    \caption{Power Usage Effectiveness (PUE) by country for our 11 selected countries. }
    \label{tab:pue}
\end{table}

We consider Microsoft's  U.S. average PUE of 1.17 \cite{Microsoft_Water_2022_how_azure_cloud} and Equinix's  
global average PUE of 1.42 \cite{equinixPUE}
for the U.S. and global averages, respectively.

\subsubsection{Energy consumption}
Exact LLM inference energy costs are often lacking in the public domain, especially for powerful 
proprietary LLMs such as GPT-4 deployed in real-world inference systems. 
To estimate LLM inference energy,
some studies resort to a commonly cited claim that
each request of the GPT model family underlying
ChatGPT consumes about 10x the energy as a Google search \cite{Datacenter_EnergyEstimate_IEA_2024}, while others
use GPUs' processing capability in tera operations per second (TOPS) and 
the power consumption reported
by manufacturers \cite{Carbon_ChatGPT_Lower_Human_Tomlinson_UCI_ScientificReports_2023_tomlinson2023carbon}. In practice, however, the actual LLM inference energy
consumption depends on a variety of factors, including
the hardware, service level objectives (SLOs), and system optimization \cite{AI_SplitWise_LLM_Inference_ChaojieZhang_Microsoft_ISCA_2024_10609649,AI_DynamoLLM_ResourceManagement_LLM_Inference_energy_UIUC_Microsoft_ChaojieZhang_2024_stojkovic2024dynamollmdesigningllminference}.

To estimate LLM inference energy consumption for a user-facing application (used by, e.g., ChatGPT),
we resort to an online calculator \cite{ecologits-calculator} 
and a recent study \cite{AI_DynamoLLM_ResourceManagement_LLM_Inference_energy_UIUC_Microsoft_ChaojieZhang_2024_stojkovic2024dynamollmdesigningllminference}. 
These two sources use different methods to calculate
the LLM inference energy consumption, which we describe as follows.

\paragraph{LLM inference energy estimates by \cite{ecologits-calculator,AI_DynamoLLM_ResourceManagement_LLM_Inference_energy_UIUC_Microsoft_ChaojieZhang_2024_stojkovic2024dynamollmdesigningllminference}}

The online calculator \cite{ecologits-calculator}
estimates the inference energy consumption by various LLMs
based on a transparent methodology. 
It first uses the energy measurements from a set of open-sourced models
(mostly on Nvidia A100 GPUs)
to fit an energy consumption curve in terms of the number
of model parameters. For mixture-of-expert model architectures such as the one commonly believed to be used by GPT-4, a range of active model parameters are considered.
The energy measurement takes into account a server's
non-GPU power attributed to the model depending on the fraction
of GPU resources the model utilizes. Nonetheless,
 \cite{ecologits-calculator} only considers the token generation phase, while neglecting
the prompt processing phase (i.e., processing user prompts to generate the first output token) which is also energy-intensive \cite{AI_SplitWise_LLM_Inference_ChaojieZhang_Microsoft_ISCA_2024_10609649,AI_PowerCharacterization_LLM_Inference_ChaojieZhang_Microsoft_ASPLOS_2024_10.1145/3620666.3651329,AI_DynamoLLM_ResourceManagement_LLM_Inference_energy_UIUC_Microsoft_ChaojieZhang_2024_stojkovic2024dynamollmdesigningllminference}. In addition, it does not consider batching and
essentially models a lightly-loaded system without request contention.
This can be viewed as a reference system used by industries, e.g., \cite{AI_SplitWise_LLM_Inference_ChaojieZhang_Microsoft_ISCA_2024_10609649,AI_PowerCharacterization_LLM_Inference_ChaojieZhang_Microsoft_ASPLOS_2024_10.1145/3620666.3651329} measure LLM inference energy and power consumption without batching as a reference value for energy and power provisioning,
while \cite{AI_DynamoLLM_ResourceManagement_LLM_Inference_energy_UIUC_Microsoft_ChaojieZhang_2024_stojkovic2024dynamollmdesigningllminference,AI_SplitWise_LLM_Inference_ChaojieZhang_Microsoft_ISCA_2024_10609649} use the latency measurement in such a reference system to set
real SLO targets.

On the other hand, \cite{AI_DynamoLLM_ResourceManagement_LLM_Inference_energy_UIUC_Microsoft_ChaojieZhang_2024_stojkovic2024dynamollmdesigningllminference} measures
the actual GPU energy consumption for LLMs on
enterprise-grade Nvidia DGX H100 servers. Its measurement also considers  ``state-of-the-practice'' optimization techniques commonly used in real systems, including batching. Importantly, it considers the prompt processing phase and representative SLO targets, which are both crucial for real-world LLM deployment. 

Energy estimates assuming a fully utilized system
without accounting for SLOs may not reflect the industry practice, since a fully-utilized system can lead
to significant SLO violations, 
which are not tolerable in real-world LLM
deployment, especially for commercial LLM applications such as
real-time conversations that have strict SLO targets to deliver good quality of experiences \cite{AI_SplitWise_LLM_Inference_ChaojieZhang_Microsoft_ISCA_2024_10609649,AI_DynamoLLM_ResourceManagement_LLM_Inference_energy_UIUC_Microsoft_ChaojieZhang_2024_stojkovic2024dynamollmdesigningllminference}. As a result, server resources for LLM inference are typically provisioned based on the peak demand to ensure SLOs are met at all times. 
In other words, the LLM inference servers may not be highly utilized under non-peak loads, resulting in a high energy consumption per request. 
For example, the Llama-2-70B inference for a medium-length request   on H100 GPUs consumes 9.4 Wh energy without batching \cite{AI_SplitWise_LLM_Inference_ChaojieZhang_Microsoft_ISCA_2024_10609649}, while 
the inference energy consumption is still 
over 4.0 Wh when batching is applied under various system loads using ``state-of-the-practice'' optimizations (the last column in
Table~II) \cite{AI_DynamoLLM_ResourceManagement_LLM_Inference_energy_UIUC_Microsoft_ChaojieZhang_2024_stojkovic2024dynamollmdesigningllminference}.

The measurement in  \cite{AI_DynamoLLM_ResourceManagement_LLM_Inference_energy_UIUC_Microsoft_ChaojieZhang_2024_stojkovic2024dynamollmdesigningllminference} only
includes the GPU energy consumption for a small set of open LLMs.  
To account for the non-GPU server energy consumption, we need to multiply the energy consumption in \cite{AI_DynamoLLM_ResourceManagement_LLM_Inference_energy_UIUC_Microsoft_ChaojieZhang_2024_stojkovic2024dynamollmdesigningllminference} by a factor of $1.5\sim2.0$ based on the server power provisioning breakdown \cite{AI_PowerCharacterization_LLM_Inference_ChaojieZhang_Microsoft_ASPLOS_2024_10.1145/3620666.3651329}. 

While \cite{ecologits-calculator}
and \cite{AI_DynamoLLM_ResourceManagement_LLM_Inference_energy_UIUC_Microsoft_ChaojieZhang_2024_stojkovic2024dynamollmdesigningllminference} use different methodologies, we note that the server-level inference energy consumption estimated by \cite{ecologits-calculator} (as of November 20, 2024)
is generally lower than that measured by \cite{AI_DynamoLLM_ResourceManagement_LLM_Inference_energy_UIUC_Microsoft_ChaojieZhang_2024_stojkovic2024dynamollmdesigningllminference} for the same model size, assuming the LLM inference system is optimized using
state-of-the-practice techniques in \cite{AI_DynamoLLM_ResourceManagement_LLM_Inference_energy_UIUC_Microsoft_ChaojieZhang_2024_stojkovic2024dynamollmdesigningllminference}.
For example, for Llama-3-70B to write a medium-length email with 250 tokens (or about 120-200 words), \cite{ecologits-calculator} estimates
the inference energy consumption as 2.62 Wh (after removing the PUE of 1.2 for data center overheads), whereas \cite{AI_DynamoLLM_ResourceManagement_LLM_Inference_energy_UIUC_Microsoft_ChaojieZhang_2024_stojkovic2024dynamollmdesigningllminference}
shows the server-level energy consumption is about 10 Wh (after
multiplying the value in the last column of Table~III by 1.6 to account for the non-GPU server energy).

This result might be surprising, as \cite{ecologits-calculator} does not consider system optimization or batching whereas \cite{AI_DynamoLLM_ResourceManagement_LLM_Inference_energy_UIUC_Microsoft_ChaojieZhang_2024_stojkovic2024dynamollmdesigningllminference} uses reasonable ``state-of-the-practice'' optimizations including batching.
Nonetheless,  \cite{ecologits-calculator} mostly uses A100 GPUs and does not consider prompt processing energy consumption, whereas
\cite{AI_DynamoLLM_ResourceManagement_LLM_Inference_energy_UIUC_Microsoft_ChaojieZhang_2024_stojkovic2024dynamollmdesigningllminference} uses
H100 GPUs (which may be more energy-consuming than A100 GPUs for LLM inference as shown by \cite{AI_SplitWise_LLM_Inference_ChaojieZhang_Microsoft_ISCA_2024_10609649}) and considers both prompt processing and token generation energy consumption. Additionally, the strict SLOs in real-world deployment prohibits the LLM inference system from being fully utilized. Thus, 
the LLM inference energy consumption estimated by \cite{ecologits-calculator} without system optimization could be even lower and still serves as a good reference point.

\begin{figure*}[!t]
    \centering
    \setcounter{subfigure}{0}
    \subfloat[S][Llama-3-70B]{
    \includegraphics[width=0.4\linewidth]{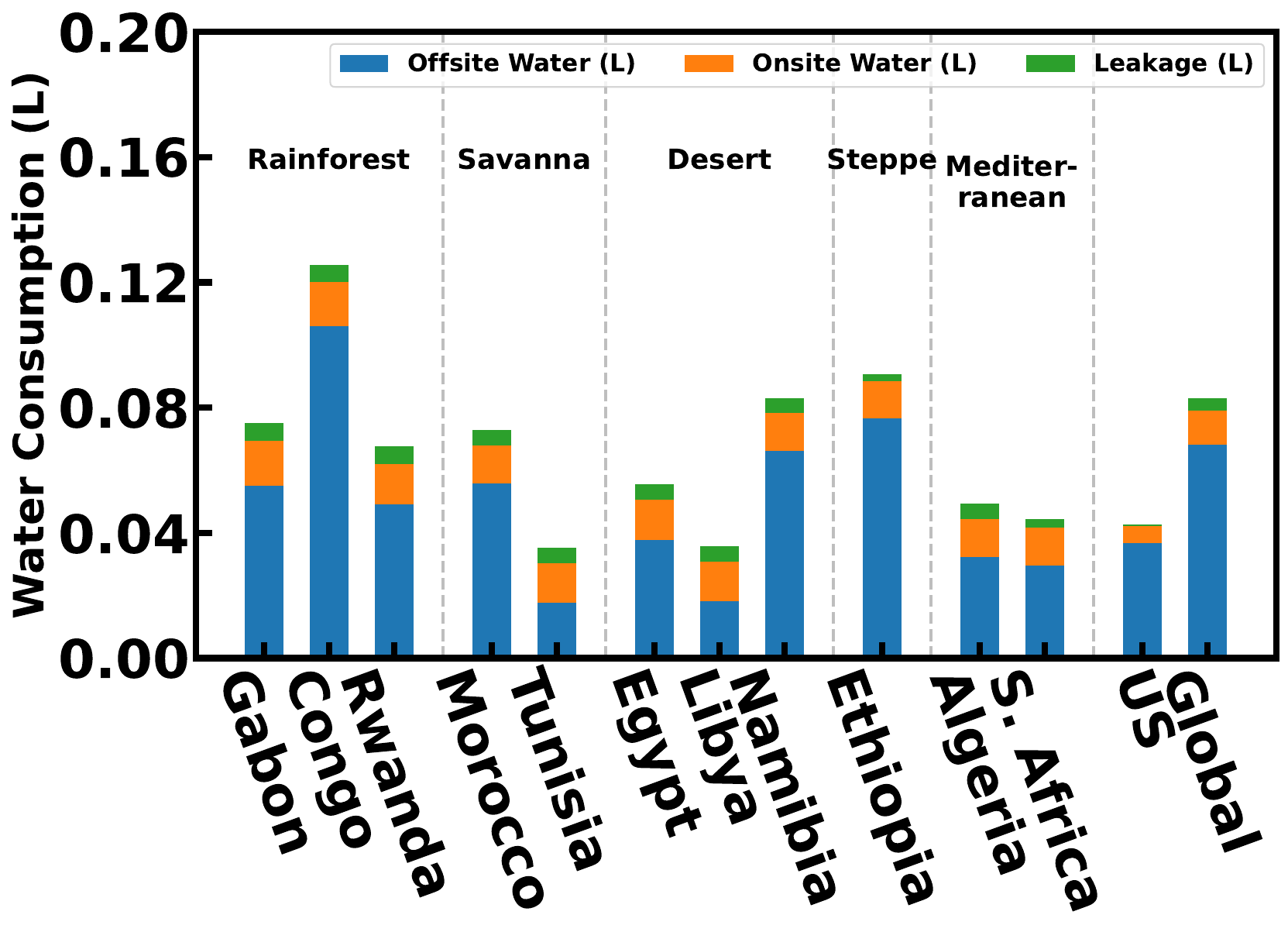}
    \label{fig:sub_llama3_8B}
    }
    \subfloat[S][GPT-4]{
    \includegraphics[width=0.392\linewidth]{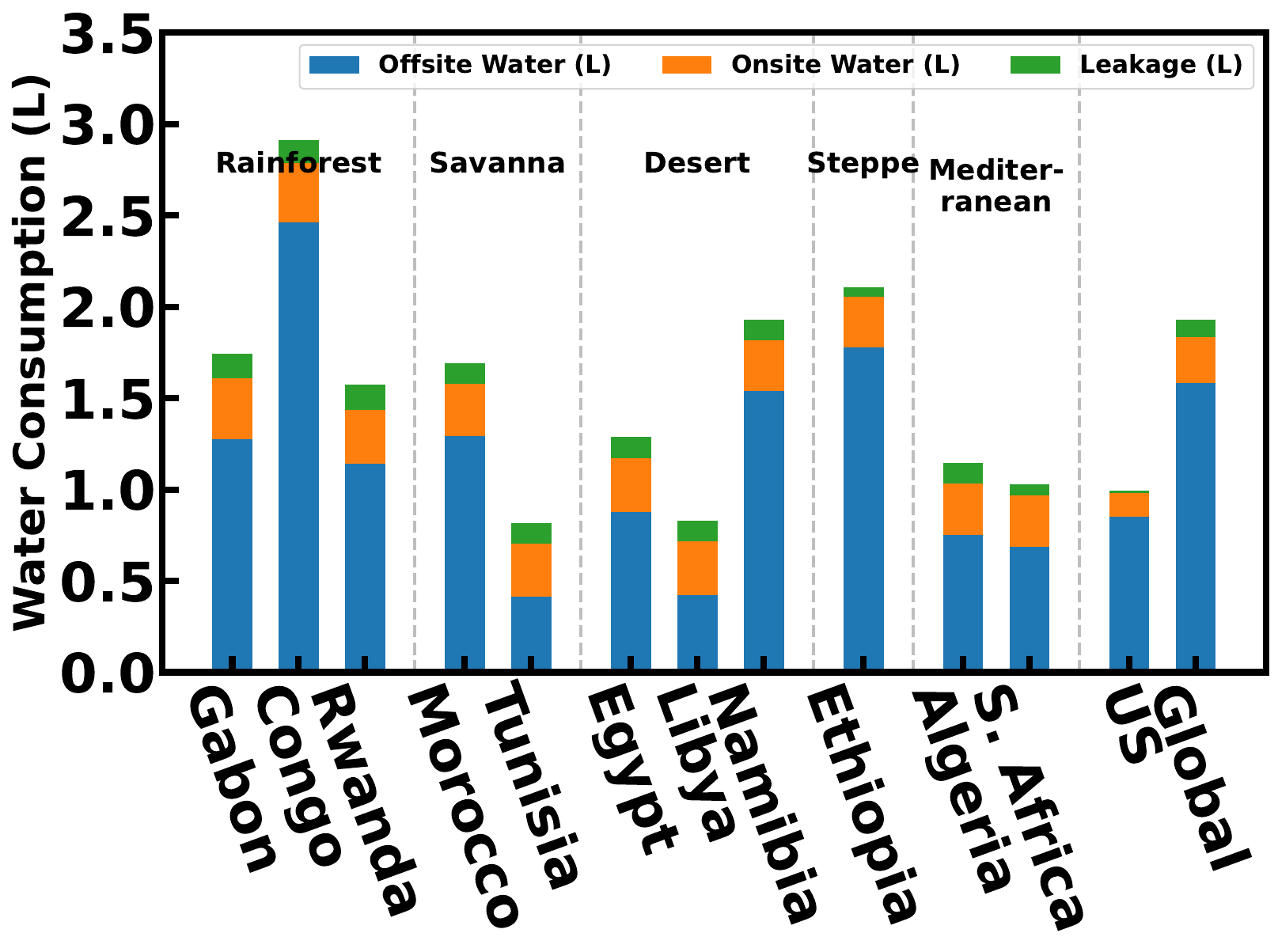}
    \label{fig:sub_GPT_4_email}
    }    \caption{Water consumption across 11 selected countries for writing a medium-length email (250 tokens) using Llama-3-70B and GPT-4, respectively. }
    \label{fig:leakage_email}
    \Description{}
\end{figure*}

\subsubsection{Energy consumption for writing a 10-page report and a medium-length email}

For the task of writing a 10-page report,\footnote{The calculator \cite{ecologits-calculator} assumes 5,000 tokens for a 5-page report. Based on the token-to-word ratio \cite{Ref27_OpenAI_Pricing_website}, we consider 5,000 tokens as roughly a 10-page report.} we assume the output is 5,000 tokens and use the estimates by \cite{ecologits-calculator}, since the energy measurement results in \cite{AI_DynamoLLM_ResourceManagement_LLM_Inference_energy_UIUC_Microsoft_ChaojieZhang_2024_stojkovic2024dynamollmdesigningllminference} do not include generating such long outputs using Llama-3-70B.
After removing the PUE of 1.2 for data center overheads,
we estimate that the energy consumption to write a 5,000-token text
by Llama-3-70B and GPT-4 are 
52.25 Wh and 4.66 kWh, respectively, based on the results
in \cite{ecologits-calculator} as of November 20, 2024. 
Note that, due to the proprietary nature of GPT-4, 
\cite{ecologits-calculator}  assumes 1,760 billion parameters for GPT-4 with a mixture-of-expert architecture based on the best-known information from various public sources. Additionally,
the energy consumption estimates for models with such large model sizes
 are based on extrapolation. As a result, without detailed information from
model owners, 
  the energy estimates for large proprietary models may have less accuracy than for small/medium open models.

For the task of writing a medium-length email, we assume the output is 250 tokens or about 120-200 words. For Llama-3-70B, by considering
a medium-length prompt and a medium system load,
we estimate
the inference energy consumption as $\sim10$ Wh after
multiplying the value in the last column of Table~III by 1.6 to account for the non-GPU server energy \cite{AI_DynamoLLM_ResourceManagement_LLM_Inference_energy_UIUC_Microsoft_ChaojieZhang_2024_stojkovic2024dynamollmdesigningllminference}.
For GPT-4, we estimate the inference energy consumption as $\sim232$ Wh \cite{ecologits-calculator}. \revise{While \cite{AI_DynamoLLM_ResourceManagement_LLM_Inference_energy_UIUC_Microsoft_ChaojieZhang_2024_stojkovic2024dynamollmdesigningllminference} provides more realistic estimates for shorter tasks by incorporating batching and prompt processing, its applicability is currently limited to one of the models we evaluate—specifically Llama-3-70B—and to shorter output tasks. For longer tasks, such as generating a 10-page report, we rely on \cite{ecologits-calculator}, which offers broader model coverage, including GPT family, albeit under simplified assumptions. Using \cite{AI_DynamoLLM_ResourceManagement_LLM_Inference_energy_UIUC_Microsoft_ChaojieZhang_2024_stojkovic2024dynamollmdesigningllminference} where available and \cite{ecologits-calculator} otherwise, our approach balances methodological rigor with practical applicability, allowing consistent estimation across diverse tasks and model architectures. }

Finally, we emphasize some limitations of our methodology. 
\revise{First, our models are themselves simplifications of the actual water usage, which has been boiled down to national levels; in reality there are substantial variations by region and time. Second, we only estimate the water usage during the model inference time, excluding the water footprint associated with idle power.}
\revise{Even under our modeling assumptions,} it is challenging to obtain precise data on the energy fuel mix and the electricity water intensity in Africa.
Moreover, the actual energy consumption of LLM inference may vary depending on the (possibly customized) optimization techniques used by real systems, particularly for the proprietary GPT-4 model. As such, our results should be regarded as first-order estimates rather than precise representations. We encourage AI model developers and data center operators to enhance transparency regarding their most recent water usage, especially in African countries facing water scarcity.

\begin{figure*}[!t]
    \centering
    \setcounter{subfigure}{0}
    \subfloat[S][Llama-3-70B]{
    \includegraphics[width=0.4\linewidth]{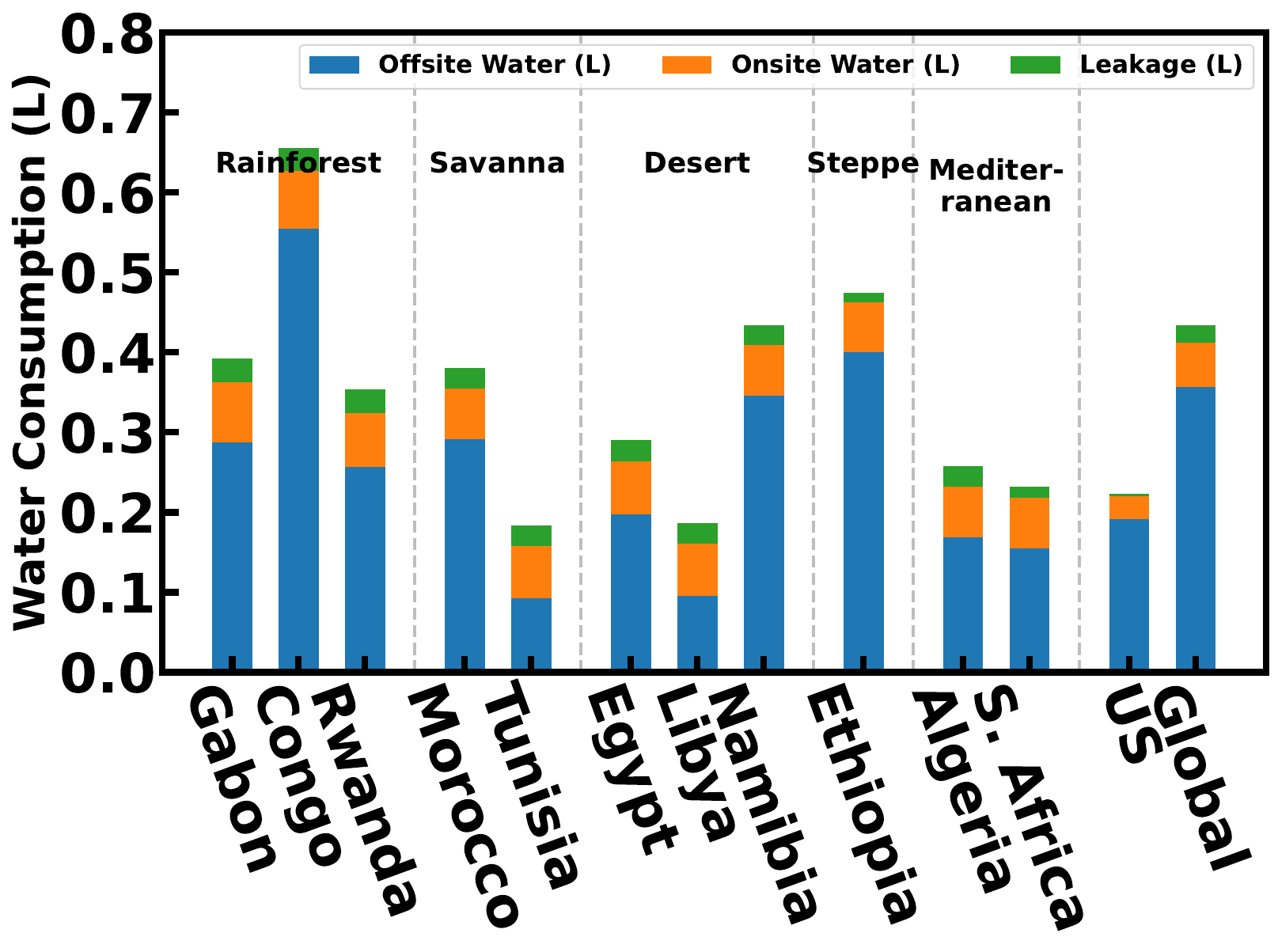}\label{fig:llama3_70B_long_report_leackage}
    }
    \subfloat[S][GPT-4]{
    \includegraphics[width=0.396\linewidth]{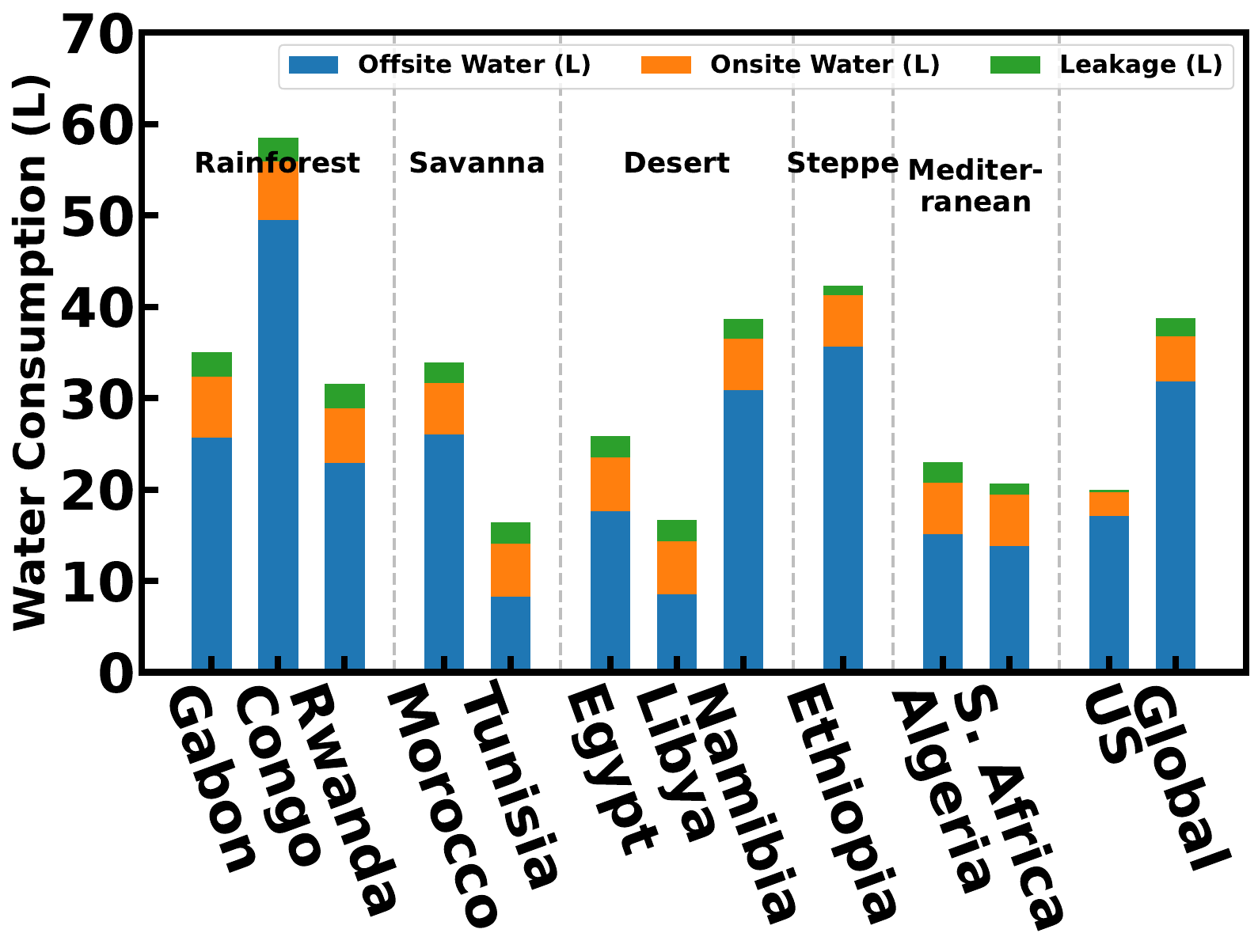}
    \label{fig:gpt-4_long_report_leakage}
    }
    \caption{Water consumption across 11 selected countries for writing a 10 page report (5000 tokens) using Llama-3-70B and GPT-4, respectively. }\label{fig:leakage_5000tokens}
    \Description{}
\end{figure*}

\subsection{Results}
To demonstrate the utility of our dataset, we use it to estimate the water consumption of two LLMs, i.e.,
Meta's Llama-3-70B and OpenAI's GPT-4, following the method in \cite{li2023making}. 
The tasks we evaluate are to write a comprehensive 10-page report and a medium-length email, as shown by Figures~\ref{fig:leakage_email}  and ~\ref{fig:leakage_5000tokens}.\footnote{\revise{We did not evaluate additional tasks because the numbers would simply scale linearly as a function of the energy consumption of each task for each model. However, the methodology would remain the same for other tasks.}} Our results indicated that writing a 10-page report using Llama-3-70B and GPT-4 in Africa could consume approximately \revise{as much as} \Noah{\textbf{0.66 liters} and \textbf{59 liters}} of water, respectively.

When comparing the water usage of these models between various African countries, the United States and the global averages, we observed the following: 
First, \Noah{9} of the 11 selected African countries have a lower water consumption than the global average. In addition, \Noah{Algeria, South Africa, and Tunisia} even have a lower water consumption than the U.S. average. This may be surprising, as Africa is commonly viewed as a water-scarce and dry continent. \Noah{The underlying reason is that the per-fuel water intensity factors ($w_k$) applicable to Africa's generation fleet are lower than the global averages underlying the Water Resource Institute  benchmark of 4.807~L/kWh \cite{reig2020guidance}. As shown in Table~\ref{tab:wk_values}, Africa-specific $w_k$ values differ from global averages for three key fuels: (i)~natural gas, where Africa's predominantly CCGT fleet \cite{siemens_egypt_ccgt,ge_tm2500_algeria} yields $w_k = 0.795$~L/kWh versus a global average of $\sim$1.58~L/kWh that includes simple-cycle plants; (ii)~hydropower, where Africa's deep, high-head reservoirs yield $w_k = 5.32$~L/kWh \cite{sanchez2020freshwater} versus global estimates of $\sim$22~L/kWh \cite{mekonnen2015water}; and (iii)~nuclear, where cooling technology differences yield 1.515 versus $\sim$2.12~L/kWh globally.

Four of our selected countries (Algeria, Tunisia, Egypt, and Libya) rely primarily on CCGT gas, which has the lowest $w_k$ among thermal fuels. Two countries (South Africa and Morocco) rely on coal ($w_k = 2.008$), which is higher than gas but still below the global weighted average. Even with higher onsite WUE and PUE values typical of African data centers \cite{DataCenter_PUE_Africa_172_2021}, the offsite advantage keeps total water consumption below the global reference for 9 of the 11 countries.}

\Noah{Second, the results in Figs.~\ref{fig:leakage_email} and~\ref{fig:leakage_5000tokens} indicate a strong relationship between the electricity fuel mix and water consumption. Ethiopia  and Republic of Congo are the only countries exceeding the global average, driven by their heavy reliance on hydroelectric power (95.7\% and 99.6\% respectively \cite{owid-energy-mix}), which has the highest water intensity due to reservoir evaporation \cite{sanchez2020freshwater,mekonnen2015water}. Countries with moderate hydro shares, such as Namibia, Rwanda, and Gabon, also approach the global average, highlighting the disproportionate impact of hydro on water usage. In contrast, gas-dominated countries (e.g., Algeria, Tunisia, and Egypt) exhibit the lowest water consumption, while coal-based systems fall in between. Overall, variation in water consumption is more strongly linked to electricity fuel mix than to climate, with hydroelectric share emerging as the primary driver.}

Regarding leakage, countries in desert regions (e.g., Libya and Egypt) exhibit higher leakage rates despite having lower overall water consumption than the global average. This highlights the role of water infrastructure inefficiencies in exacerbating water losses.
Regions such as the Steppe and Savanna have elevated leakage rates relative to their nominal onsite water consumption, suggesting that infrastructure improvements in these areas could lead to substantial water savings. More broadly, the comparison between African countries, global averages, and developed regions like the United States reveals stark disparities in leakage levels.

\begin{figure}
\centering
   \includegraphics[width=0.85\linewidth]{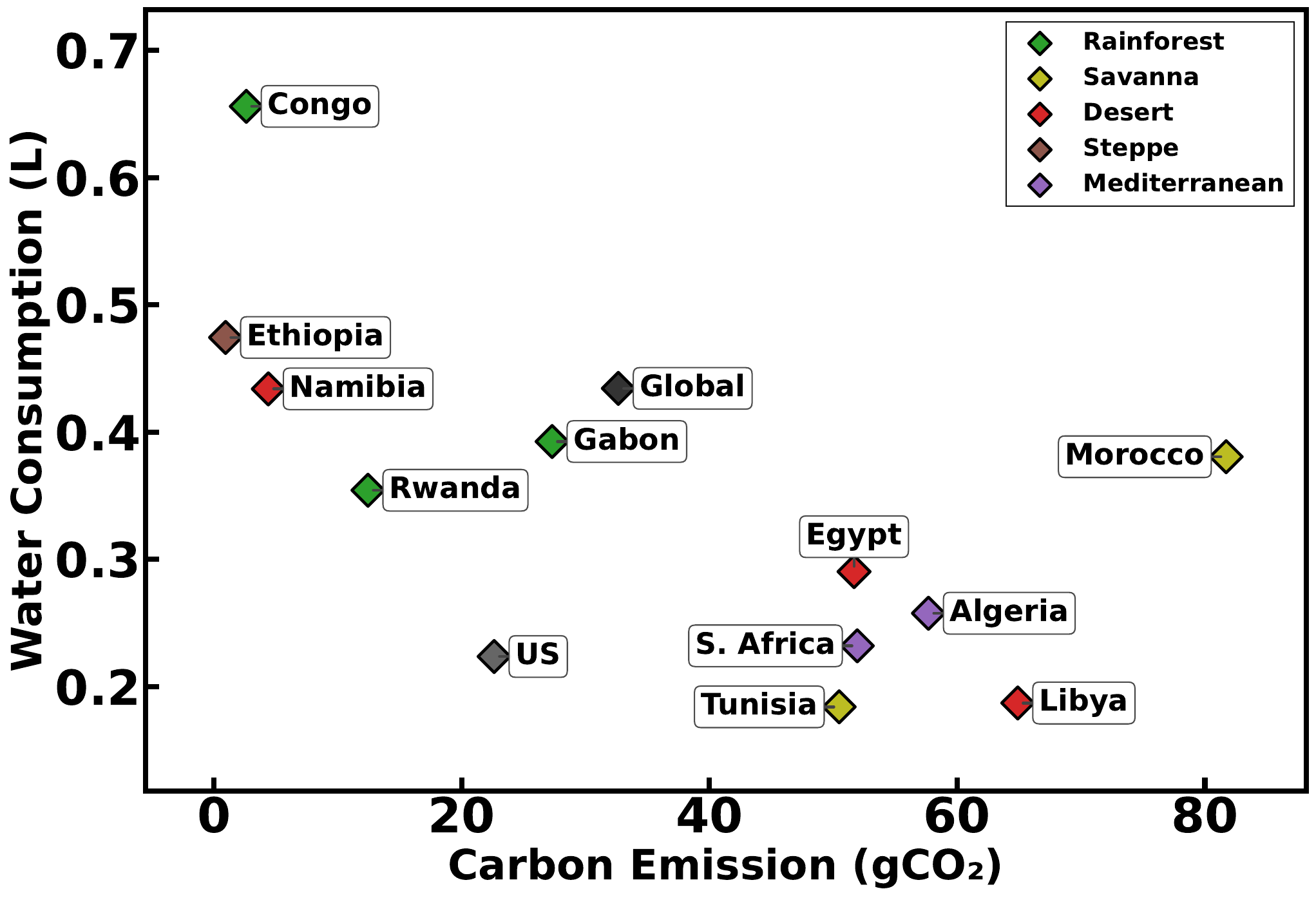}
   \caption{Water consumption and (scope-2) carbon emission across various African countries for writing a 10-page report using the Llama-3-70B model. ``Congo'' indicates Republic of the Congo.}\label{fig:water_carbon_llama_3}
   \Description{}
\end{figure}

\subsubsection{Correlation between Water Consumption and Carbon Emissions}
\label{sec:water-carbon}
Building on our findings, we explored the relationship between water consumption and carbon emissions for the Llama-3-70B model for the task of writing a 10-page report. From Figure~\ref{fig:water_carbon_llama_3}, we observe a tradeoff between water consumption and carbon emission, which is consistent with the findings in prior studies \cite{Shaolei_Water_SpatioTemporal_GLB_TCC_2018_7420641}.
This prompts further attention to strike a balance between water consumption and carbon emissions to enable truly sustainable AI in African countries.

\begin{figure*}[!t]
    \centering
    \setcounter{subfigure}{0}
    \subfloat[S][Llama-3-70B]{
    \includegraphics[width=0.45\linewidth]{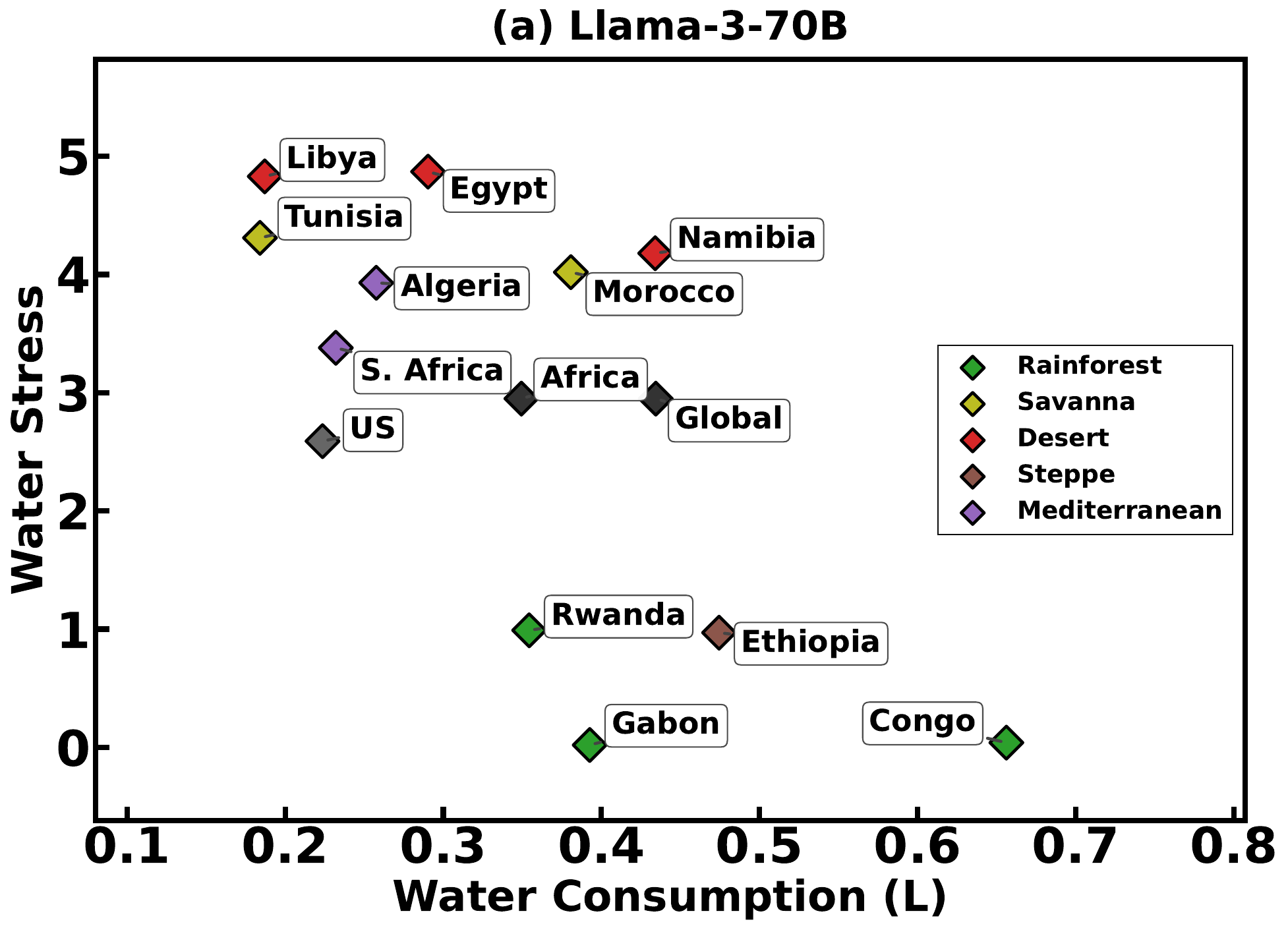}\label{fig:llama3_70B_long_report_Water Stress}
    }
    \quad
    \subfloat[S][GPT-4]{
    \includegraphics[width=0.45\linewidth]{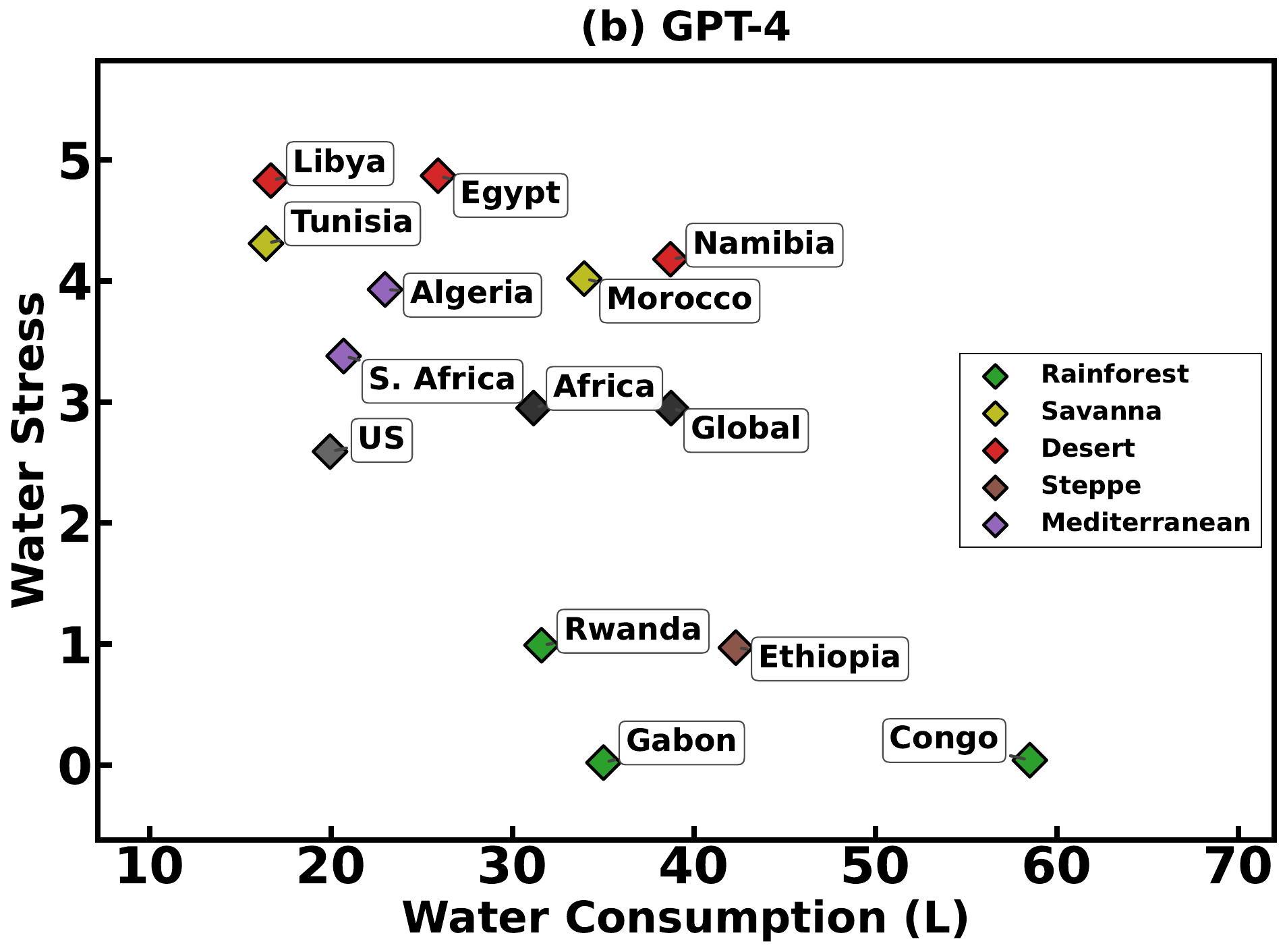}
    \label{fig:gpt-4_long_WaterStress}
    }
        \caption{Water consumption and Water Stress levels across 11 selected countries for writing a 10 page report (5000 tokens) using Llama-3-70B and GPT-4, respectively. 
        ``Africa" indicates the African average, ``Congo'' indicates Republic of the Congo and ``Global" indicates the global average.}\label{fig:Water_Stress_11_Countries}
        \Description{}
\end{figure*}
\subsubsection{Incorporating Water Stress Levels}

Figure 5 presents a comparison of total water consumption against the existing water stress levels in the 11 selected countries, analyzing the potential impact of placing data centers in these regions.\Noah{Ethiopia and the Republic of the Congo—the two countries above the global average—both exhibit low water stress levels (Baseline Water Stress (BWS) $\sim$<$1$), suggesting that their relatively high water consumption from hydroelectric generation is unlikely to pose a significant threat to local water availability. Rwanda and Gabon, which sit just below the global average due to moderate hydro shares, also have low water stress. On the other hand, several North African countries (Egypt, Libya, Tunisia) face extremely high water stress (BWS~$>$~4) despite having relatively low AI water consumption due to their gas-dominated grids. A notable case is Namibia, where water consumption is near the global average and water stress is very high (BWS~4.18); deploying water-intensive AI models there could place additional strain on an already scarce resource.}

\begin{figure*}[!t]
    \centering
    \setcounter{subfigure}{0}
    \subfloat[S][Llama-3-70B]{
    \includegraphics[width=0.45\linewidth]{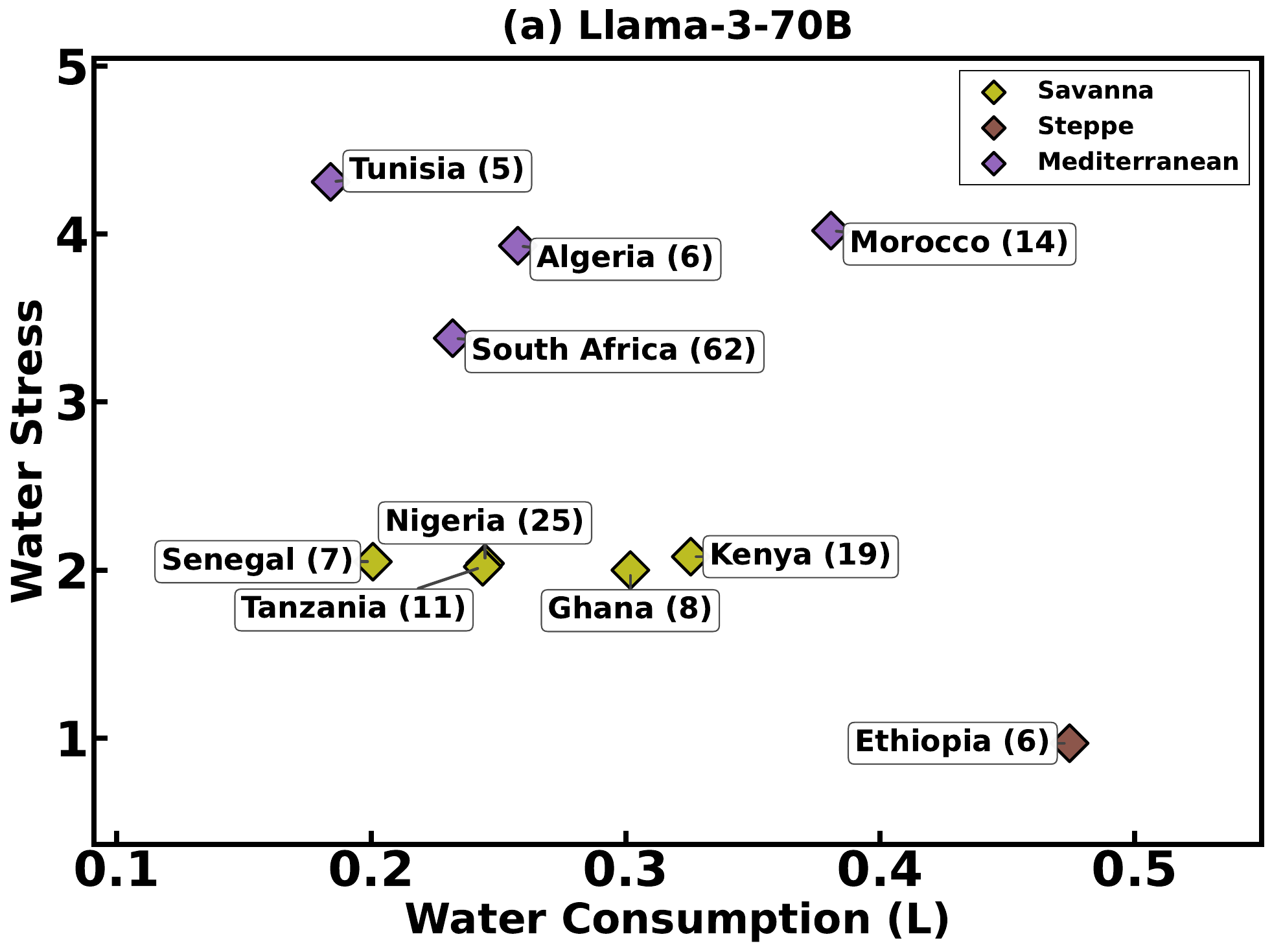}\label{fig:llama3_Top10Countries}
    }
    \quad
    \subfloat[S][GPT-4]{\includegraphics[width=0.45\linewidth]{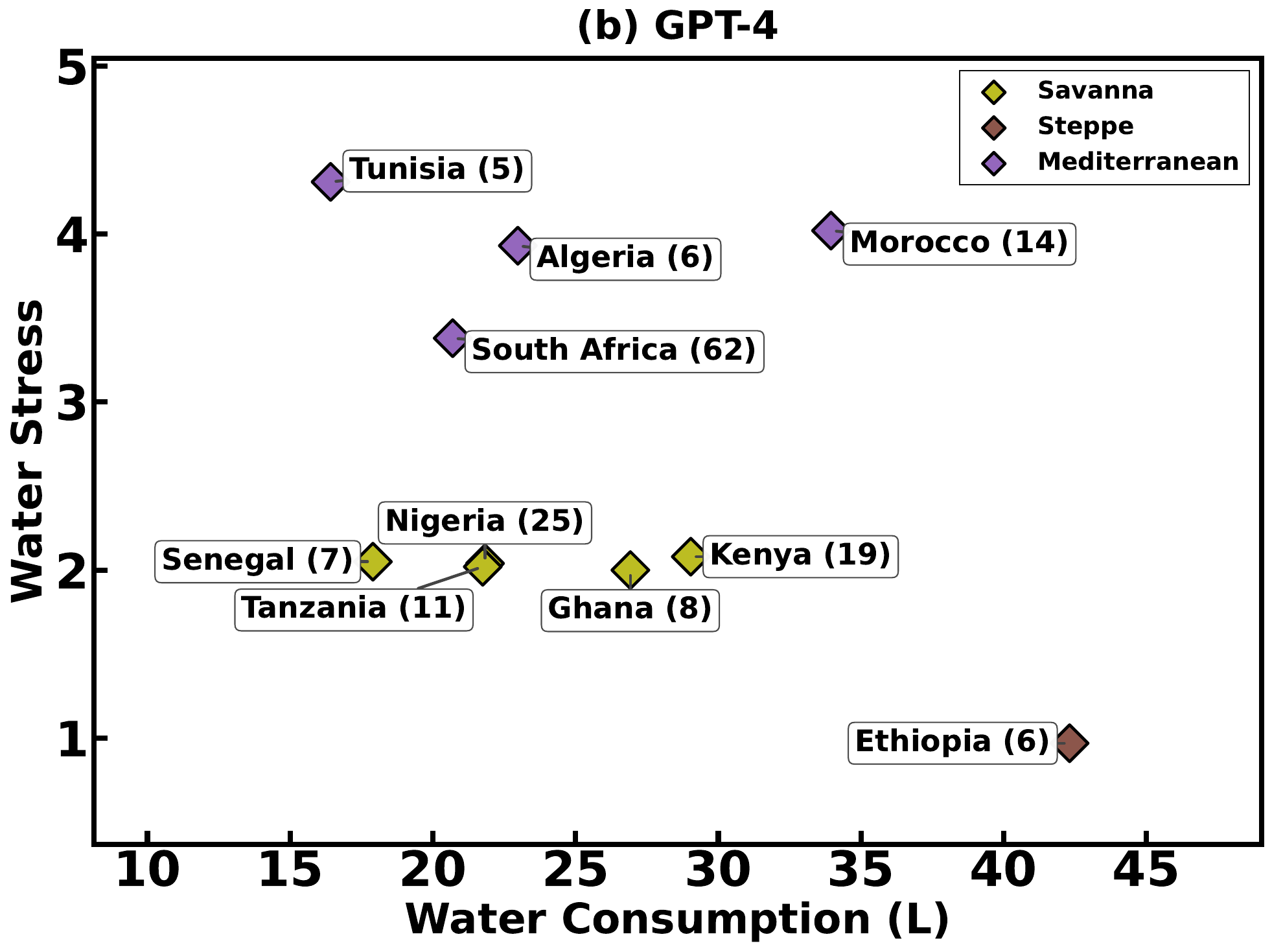}
    \label{fig:gpt-4_Top10Countries}
    }
    \label{fig:sub_GPT_4_dc}
        \caption{Water Consumption and Stress Levels in African Countries with high data center density (number of data centers in parenthesis), evaluated for writing a 10 page report (5000 tokens) using Llama-3-70B and GPT-4, respectively.}\label{fig:Top10Countries}
        \Description{}
\end{figure*}
To specifically examine the impact of data centers on local water resources, we then narrowed our focus to the  countries in Africa with the most existing data centers \cite{datacenter_map_dataset}. Figure~\ref{fig:Top10Countries} shows a trend where countries such as South Africa and Nigeria, which have numerous data centers, also exhibit high water consumption \revise{per model inference}. This could be indicative of the greater strain these data centers place on local water resources, especially in regions with high levels of water stress. Our data suggests that without strategic intervention, the expansion of data centers could exacerbate local water scarcity issues, placing additional strain on already limited water resources. Strategic placement should prioritize areas with lower water stress levels or where advanced cooling technologies can be implemented effectively. For future placement, the ideal "sweet spot" for situating data centers in Africa would be regions that demonstrate a lower water stress level coupled with moderate water consumption. This criterion suggests that data centers could be strategically placed in areas where they would exert a minimal impact on local water resources while still fulfilling operational demands.

\section{Conclusion}
This study presents the first-of-its kind large-scale evaluation of water efficiency for African data centers, providing a crucial dataset that accounts for both direct and indirect water consumption across diverse climatic regions. Our findings highlight significant regional variations, demonstrating that \Noah{most African countries (9 of 11 selected) exhibit lower AI-related water consumption than global averages, primarily because they generate electricity from fuels with low water intensities. However, countries with hydro-dominated electricity grids---such as Ethiopia and Republic of Congo---face substantially higher water consumption due to the water-intensive nature of hydroelectric generation. Our analysis suggests that the electricity fuel mix is a stronger predictor of cross-country variation than climate region alone, which has practical implications for data center site selection and infrastructure planning.}
This study underscores the need for site-specific infrastructure planning and adaptive cooling technologies to mitigate these challenges.

However, limitations such as the aggregated nature of water stress data and variations in cooling technologies indicate that further refinements are needed to enhance the accuracy of AI-related water impact assessments. Moving forward, policymakers, researchers, and data center operators can prioritize sustainable water management strategies by improving infrastructure to minimize leakage, adopting water-efficient cooling systems, and ensuring AI deployment aligns with local water availability.

\section*{Acknowledgments}
\revise{
This work was made possible in part by funding from the Gates Foundation. 
The views and opinions expressed in this study are those
of the authors and do not necessarily reflect the views or
positions of the sponsors.
}

\clearpage

\bibliographystyle{ACM-Reference-Format}

\nocite{*}

\bibliography{cited.bbl}

\end{document}